%% file: acl_latex.tex
\pdfoutput=1

\documentclass[11pt]{article}

\usepackage[preprint]{acl}

\usepackage{times}
\usepackage{latexsym}
\usepackage{amsmath}

\usepackage[T1]{fontenc}

\usepackage[utf8]{inputenc}

\usepackage{microtype}

\usepackage{inconsolata}

\usepackage{graphicx}
\usepackage{booktabs}
\usepackage{amsfonts}

\input{packages}

\title{$I^2G$: Generating Instructional Illustrations via Text-Conditioned Diffusion}

\author{
  \textbf{Jing Bi\textsuperscript{1}},
  \textbf{Pinxin Liu\textsuperscript{1}},
  \textbf{Ali Vosoughi\textsuperscript{1}},
  \textbf{Jiarui Wu\textsuperscript{1}},
\\
  \textbf{Jinxi He\textsuperscript{1}},
  \textbf{Chenliang Xu\textsuperscript{1}}
\\
\\
  \textsuperscript{1}University of Rochester
}

\begin{document}
\maketitle
\begin{abstract}
The effective communication of procedural knowledge remains a significant challenge in natural language processing (NLP), as purely textual instructions often fail to convey complex physical actions and spatial relationships. We address this limitation by proposing a language-driven framework that translates procedural text into coherent visual instructions. Our approach models the linguistic structure of instructional content by decomposing it into goal statements and sequential steps, then conditioning visual generation on these linguistic elements. We introduce three key innovations: (1) a constituency parser-based text encoding mechanism that preserves semantic completeness even with lengthy instructions, (2) a pairwise discourse coherence model that maintains consistency across instruction sequences, and (3) a novel evaluation protocol specifically designed for procedural language-to-image alignment. Our experiments across three instructional datasets (HTStep, CaptainCook4D, and WikiAll) demonstrate that our method significantly outperforms existing baselines in generating visuals that accurately reflect the linguistic content and sequential nature of instructions. This work contributes to the growing body of research on grounding procedural language in visual content, with applications spanning education, task guidance, and multimodal language understanding.
\end{abstract}

\section{Introduction}
Procedural language understanding represents a significant challenge in natural language processing. Unlike declarative text, instructional language contains implicit temporal dependencies, causal relationships, and action sequences that require specialized approaches to model effectively. While large language models have advanced our ability to generate and comprehend procedural text, they still struggle with the inherent limitations of the text modality itself, which often inadequately conveys spatial relationships, physical manipulations, and visual states critical to executing complex tasks. Users frequently supplement textual instructions with visual aids, demonstrating the natural complementary relationship between language and vision in procedural understanding. This paper addresses this gap by developing a framework that bridges procedural language with corresponding visual representations, advancing multimodal instruction understanding in NLP.
\begin{figure}[t]
    \centering
    \includegraphics[width=1.0\linewidth]{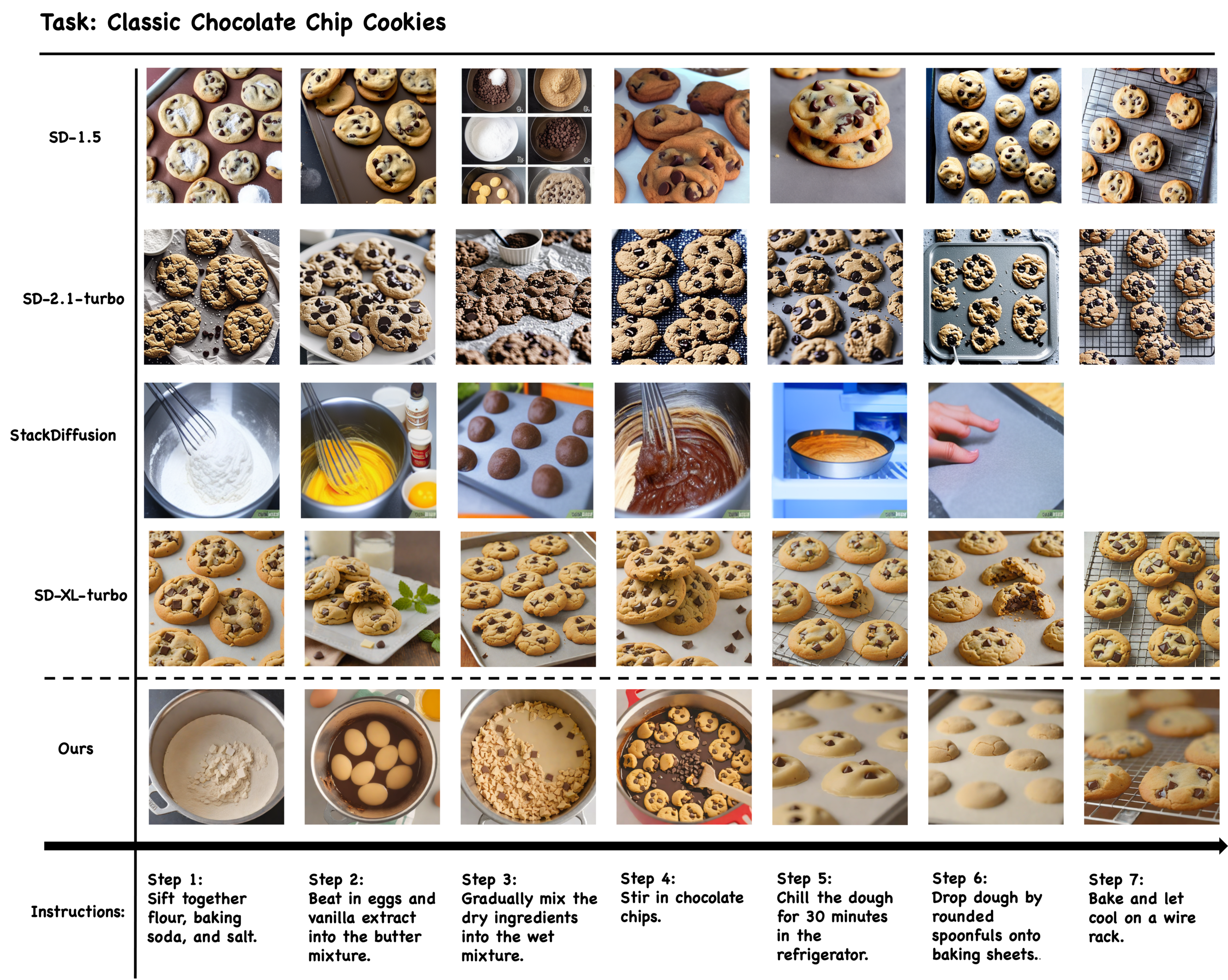}

    \caption{The qualitative results of our method, compared with baseline models, are illustrated in the figure. The baseline model struggles to capture the progression in the text, whereas our method successfully captures this progression and achieves a more illustrative result. However, the image lacks in StackDiffusion due to the model's limitation of not generating more than 6 steps.} 
    \label{fig:instruct}
    \vspace{-5mm}
\end{figure}

To address this need for visual instruction, platforms such as YouTube and TikTok have become go-to resources for learning new skills. Instructional videos on these platforms have emerged as a popular means for people to learn new skills and improve their abilities in executing complex procedural activities. Typically, an instructional video includes a brief title that states the ultimate goal to be achieved. Throughout the video, various steps are demonstrated progressively, each building on the dependencies of the previous ones. In key time frames, visual instructions are supplemented with specific textual descriptions of the actions or content of the current step. This combination allows viewers to grasp the nuances of complex tasks by linking visual demonstrations with textual explanations. These videos are not only beneficial for humans, but also hold promise for machine learning, as they provide clear visual demonstrations of intricate tasks and detailed human-object interactions across various domains. These challenges intersect with core NLP tasks such as procedural text understanding and multimodal instruction following, which have gained interest in the ACL community.

Drawing inspiration from these instructional videos, we formulate this process as a generative problem. The challenge lies in effectively modeling the distribution of the goals of a task, its procedural steps, and the associated visual information. Our approach addresses this by utilizing a generative model to encapsulate instructional visual information. We decompose the problem into manageable parts: modeling the textual components of goals and steps, and then conditioning the visual information on these textual descriptors. This is mathematically represented through the probability function
\begin{equation}
p(g,s,v) = p_{\phi}(\mathcal{V}| g,\mathcal{S}) p_{\theta}(g,\mathcal{S}) ,
\end{equation}
where \( p(\cdot) \) denotes the model distribution, and \( g \), \( \mathcal{S} \), and \( \mathcal{S} \) represent the goal text, the sequence of step instructions $(s_i \in \mathcal{S})$, and the visual information (such as images or videos)  $(v_i \in \mathcal{V})$, respectively.

To effectively learn \( p_{\phi}(v | g,s) \), we employ a diffusion-based strategy, specifically training a denoising diffusion model. For the \( p_{\theta}(g,s) \) distribution, we first leverage several recent instructional video and image datasets containing trajectory data of task performance, ensuring our model learns a wide array of task-specific visuals that are both accurate and varied. To enhance the integration and coherence between the textual and visual components, we refine our approach by leveraging a large language model (LLM) extensively trained to represent the joint distribution of goals and procedural steps \( p_{\theta}(g,s) \), generating text samples that are diverse and contextually relevant.
We leverage the pretrained LLM to draw $(g,s)$ samples and use image-text alignment model acts as a reward model to provide feedback to further ensure that the generated visuals accurately reflect the textual instructions.

Evaluating the proposed illustrative instruction generation framework presents unique challenges, especially given the limitations of current mainstream evaluation metrics for instructional content. Effective evaluation requires considering both the quality of the generated images and their alignment with the instructional text. Traditional metrics like Fréchet Inception Distance (FID) and Inception Score focus on photorealism but fail to capture the semantic alignment between text and images.

CLIPScore is commonly used for evaluating text-image alignment, as it measures how well text descriptions match generated images. However, our experiments revealed that CLIPScore often falls short in capturing the nuances of longer, descriptive texts typical of procedural steps, leading to misleadingly low alignment scores. State-of-the-art multimodal models, which use techniques like language binding or image binding, also struggle to provide meaningful judgments on the congruence between extended texts and corresponding images.

To address these evaluation challenges, we initially conducted a series of human evaluations to assess the alignment between the generated text and images. Human judgment remains the gold standard for measuring the quality and relevance of instructional content, providing insights that automated metrics currently cannot. Based on these human evaluations, we then leveraged caption generation and text similarity models that demonstrated high alignment with human judgments. These models were adapted to refine our evaluation framework, ensuring that it not only measures photorealism but also accurately reflects the instructional integrity and coherence between text and visual representations.

Our main contributions are summarized as follows:

\begin{itemize}
    \item We propose a mathematically sound framework for visual instruction generation based on goal and step dependencies. This framework effectively decomposes the problem into manageable components, enabling meaningful generation of instructional content.
    \item We design a feedback-based finetuning mechanism for a diffusion generative model. By incorporating image-text alignment feedback, we ensure that the visual outputs are relevant and accurate representations of the textual instructions.
    \item We address the challenges in evaluating visual instruction generation by developing a comprehensive evaluation protocol. To our knowledge, we are the first to conduct experiments targeting instructional image-text alignment within the LLM realm. This protocol includes human evaluations and advanced models for caption generation and text similarity, ensuring robust assessment of the quality and coherence of the generated content.
    \item We situate our work in the broader context of multimodal text generation, demonstrating how large language models (LLMs) and diffusion can jointly produce coherent instructional visuals.
\end{itemize}

We believe that the release of our code and backbone network weights will facilitate further research and benchmarking in this area, driving advancements in the integration of visual and textual instructional systems.


\section{Related Work}
\label{Related Work}

\noindent \textbf{Instructional Visual Understanding:} The field of instructional visual understanding has rapidly advanced with the introduction of specialized datasets like Breakfast~\cite{Breakfast}, 50Salads~\cite{50Salads}, COIN~\cite{COIN}, and CrossTask~\cite{zhukov2019crosstask}, enabling detailed research into the structured sequences of actions in videos. The HT-Step dataset~\cite{afouras2023htstep}, in particular, with its detailed annotations from real-world instructional articles focused on cooking, provides a comprehensive framework for training models on video-text synchronization. This enhances our ability to interpret procedural videos through computational tasks such as step classification~\cite{stepclass}, action segmentation~\cite{actionseg}, video retrieval~\cite{videoretr}, and temporal article grounding~\cite{temporal}. These tasks are crucial for the precise identification and categorization of actions, even when steps are not visually explicit. Additionally, the Visual Goal-Step Inference (VGSI) task advances this domain by testing models' abilities to discern the intent behind actions and connect these with instructional goals~\cite{yang2021visual}. It enriched the interpretative capabilities of instructional content within images contexts. The introduction of the CaptainCook4D dataset further enriches this domain ~\cite{peddi2023captaincook4d}, offering recordings of cooking procedures in real kitchen environments to benchmark understanding of procedural activities visually and temporally. The availability of extensive video-text datasets has driven innovations in joint video-language embedding ~\cite{bain2022frozen, miech2020endtoend, radford2021learning, yang2021taco}, particularly by utilizing narrations to contextually ground procedural steps ~\cite{han2022temporal}, enhancing performance across tasks. New benchmarks such as \emph{InstructionBench}~\cite{wei2025instructionbench} explicitly evaluate question-answering over instructional videos, while \emph{Pivot} pre-training exploits task hierarchies for more data-efficient representation learning~\cite{samel2025pivot}.  Cross-modal planning has been pushed forward by \emph{PlanLLM}~\cite{yang2025planllm}, and large-scale transfer of internet video knowledge to unseen tasks is demonstrated by Luo \textit{et al.}~\cite{luo2025internetvideo}.  

\noindent \textbf{Aligning Diffusion Models:} A variety of works attempt to improve diffusion models for aligning with human preferences. Many approaches focus on fine-tune for a better alignment. DDPO and DPOK~\cite{black2024training, fan2023dpok} use reinforcement learning approaches for reward fine-tuning of diffusion models. With a slight difference in approach, a work~\cite{lee2023aligning} uses binary human feedback and semi-supervised learning for the reward fine-tuning. DRaFT, AlignProp, and ImageReward~\cite{clark2023directly, prabhudesai2023aligning, xu2023imagereward} directly use reward function gradients to fine-tune diffusion models. Instead of using a reward function, Diffusion-KTO, Diffusion-DPO, and D3PO~\cite{Li2024AligningDM, wallace2023diffusion, Yang2023UsingHF} introduce a fine-tuning objective directly targeting human preferences and feedback for alignment with human preferences. Emu~\cite{dai2023emu} selects an extremely small number of high-quality data in the process of fine-tune for better generated image quality of text-to-image models. Similarly, RAFT~\cite{dong2023raft} samples and filters high-reward data based on a reward model and uses the filtered data to fine-tune diffusion models. PickScore and Human Preference Score v2~\cite{kirstain2023pickapic, wu2023human} work as scoring models to guide the fine-tuning of diffusion models, better aligning them with human preferences. 

Some works propose techniques or models which lead to a better human preference alignment. SDXL~\cite{podell2023sdxl} improves diffusion models by adding a larger UNet-backbone, an additional conditioning technique and a noising-denoising process based on the existing latent diffusion model. Several methods~\cite{gal2022image, hao2023optimizing} optimize conditional input prompts in text-to-image models. Some other methods~\cite{BetkerImprovingIG, segalis2023picture} use captions with better fidelity and semantics to recaption existing datasets for improved text-to-image generation quality. \emph{Calibrated Preference Optimization (CaPO)} introduces a multi-reward alignment strategy for text-to-image diffusion~\cite{lee2025capo}, while \emph{VideoDPO} brings omni-preference alignment to text-to-video generation~\cite{liu2025videodpo}.  A comprehensive 2025 survey by Wu \textit{et al.}~\cite{wu2025survey} synthesizes these emerging directions.

\noindent \textbf{Language Models as Evaluators:} Recognizing the burgeoning capabilities of LLMs for open-ended tasks, evaluation methodologies have shifted towards a direct appraisal of generated text quality, often employing LLMs themselves as evaluators~\cite{fu2023gptscore, wang2023chatgpt, wang2023large}. The automation of evaluation is further extended to the multi-modality domain~\cite{chen2024mllmasajudge}. X-IQE~\cite{chen2023xiqe} leverages a hierarchical chain of thoughts (CoT), from image description to task-specific analysis for reasoning the Text2Image generation quality. Concept coverage~\cite{chen2024evaluating} calculated based on Visual Question Answering semantic entropy is further proposed for aestheticism and defects qualification of generations. \emph{CIGEval} frames GPT-4-based vision-language agents as fine-grained judges for conditional image generation~\cite{wang2025cigeval}, and \emph{GPT-ImgEval} provides the first dedicated benchmark for diagnosing GPT-4V’s image generation and editing abilities~\cite{yan2025gptimgeval}.

\section{Preliminary}
\subsection{Generative Diffusion Models}
\begin{equation}
    p_{\phi}(z_{0:T}) := p(z_T) \prod_{t=1}^{T} p_{\phi}(z_{t-1}|z_t), 
\end{equation}
where $p_{\phi}$ is a model distribution parameterized by $\phi$ and $z_1, \dots, z_T$ are latent variables of the same dimensionality as $z_0$. Conversely, the forward process models $q(z_{1:T}|z_0)$ by gradually adding Gaussian noise to the data sample $z_0$. In this process, the intermediate noisy sample $z_t$ can be sampled as:
\begin{equation}
    z_t = \sqrt{\alpha_t}z_0 + \sqrt{1 - \alpha_t}\epsilon, 
\end{equation}
in variance-preserving diffusion formulation ~\cite{ho2020denoising}. Here, $\epsilon \sim \mathcal{N}(0, I)$ is a noise variable and $\alpha_{1:T} \in (0, 1]^T$ is a sequence that controls the amount of noise added at each diffusion time $t$. Given the noisy sample $z_t$ and $t$, the diffusion model $f_{\phi}$ learns to approximate the reverse process for data generation. The diffusion model parameters $\phi$ are typically optimized to minimize $\mathbb{E}_{z_t,\epsilon} \| \epsilon - f_{\phi}(z_t, t) \|_2$  or $\mathbb{E}_{z_t,\epsilon} \| z_0 - f_{\phi}(z_t, t) \|_2$. Note that exact formulations vary across the literature, and we kindly refer the reader to the survey papers ~\cite{chang2023design, yang2024diffusion} for a more comprehensive review of diffusion models.

\subsection{Classifier-Free Guidance (CFG)}

CFG is a method proposed to achieve a better trade-off between fidelity and diversity for conditional sampling using diffusion models. Instead of generating a sample using conditional score estimates only, it proposes to mix the conditional and unconditional score estimates to control a trade-off between sample fidelity and diversity:
\begin{equation}
    \tilde{f}_{\phi}(z_t, t, c) = (1 + w)f_{\phi}(z_t, t, c) - wf_{\phi}(z_t, t, \emptyset),
\end{equation}
where $c$ is conditioning information and $w$ is a hyperparameter that controls the strength of the guidance. However, Equation 4 requires training both conditional and unconditional diffusion models. To address this, Ho et al. ~\cite{ho2020denoising} introduces conditioning dropout during training, which enables the parameterization of both conditional and unconditional models using a single diffusion network. Conditioning dropout simply sets $c$ to a null token $\emptyset$ with a chosen probability $p_{\text{uncond}}$ to jointly learn the conditional and unconditional scores during network training. Due to its ability to achieve a better balance between fidelity and diversity, CFG is used in many state-of-the-art conditional diffusion models.

\section{Method}
Our framework addresses the fundamental NLP challenge of translating procedural language into coherent visual sequences. This task requires deep understanding of linguistic structures that encode temporal relationships, action sequences, and state changes - elements that are central to procedural text understanding. We approach this as a language-conditioned generation problem with particular attention to preserving discourse coherence across sequential instructions.

\subsection{Formulation}
We aim to generate a sequence of images $\mathcal{V}=\{v_i\}$ given a goal text $g$ and step texts $\mathcal{S}=\{s_i\}$. Formally:
\[
p(\mathcal{V}\mid g,\mathcal{S}) \;=\; p\bigl(v_1,\ldots,v_n\mid g,s_1,\ldots,s_n\bigr).
\]
A naive approach assumes each $v_i$ depends only on $(g,s_i)$:
\[
p\bigl(v_1,\ldots,v_n\mid g,\mathcal{S}\bigr)\approx\prod_{i=1}^{n}p\bigl(v_i\mid g,s_i\bigr),
\]
but often fails to maintain coherence across multiple steps (see Figure~\ref{fig:instruct}). Although a fully joint model \cite{menon2023generating} better captures global context, it is fixed in step count and grows computationally expensive. We therefore introduce a \emph{pairwise} factorization:
\[
p\bigl(v_1,\ldots,v_n\mid g,\mathcal{S}\bigr)\approx\prod_{i<j}p\bigl(v_i,v_j \mid g,s_i,s_j\bigr),
\]
allowing localized interactions across adjacent steps without incurring a combinatorial explosion. We implement this with Stable Diffusion XL to learn $p\bigl(v_i,v_j\mid g,s_i,s_j\bigr)$, striking a balance between capturing relevant cross-step dependencies and maintaining efficiency.

Notably, our formulation views instruction generation as a sequence of overlapping pairs that collectively encode the trajectory. This implicitly enforces shared elements (e.g., recurring objects) while allowing each pair to focus on the step-specific text. Such a decomposition is particularly useful in real-world tasks like cooking or assembly, where continuous reference to previously introduced items or partial progress is required.

\subsection{Enhanced Cross-Image Consistency}
\label{sec:cross}
Given pairs $(v_i,s_i)$ and $(v_j,s_j)$, we first encode each image into latent tensors $z_i,z_j\!\in\!\mathbb{R}^{w\times h\times d}$ via a VAE. We assemble $Z^{\top}=[z_i^{\top},z_j^{\top}]\!\in\!\mathbb{R}^{(2m)\times d}$ where $m=w\!\times\!h$. Inspired by self-attention, we create a mask $\hat{M}\in\mathbb{R}^{(2m)\times(2m)}$ that selectively highlights temporal neighbors:
\[
\text{Softmax}\Bigl(\hat{M}\odot\bigl(\tfrac{QK^\top}{\sqrt{d}}\bigr)\Bigr),
\]
where $Q,K,V$ are projections of $Z^{\top}$. Unlike standard transformers, this mask enforces step ordering by restricting attention mostly to one’s own latent slice and the adjacent slice. Figure~\ref{fig:goal-align} illustrates the idea: each latent can still glimpse cross-step features, but is guided toward local consistency.

This design is motivated by the fact that many procedural tasks exhibit visual continuity—e.g., the same tool or object might appear in subsequent steps. Our modified attention ensures that the generated image for step $j$ inherits relevant context from step $i$, such as object identity or environment setting, yielding smoother transitions between steps without explicitly encoding an entire multi-step prompt at once.

\begin{figure*}[t!]
  \centering
  \includegraphics[width=1.0\textwidth]{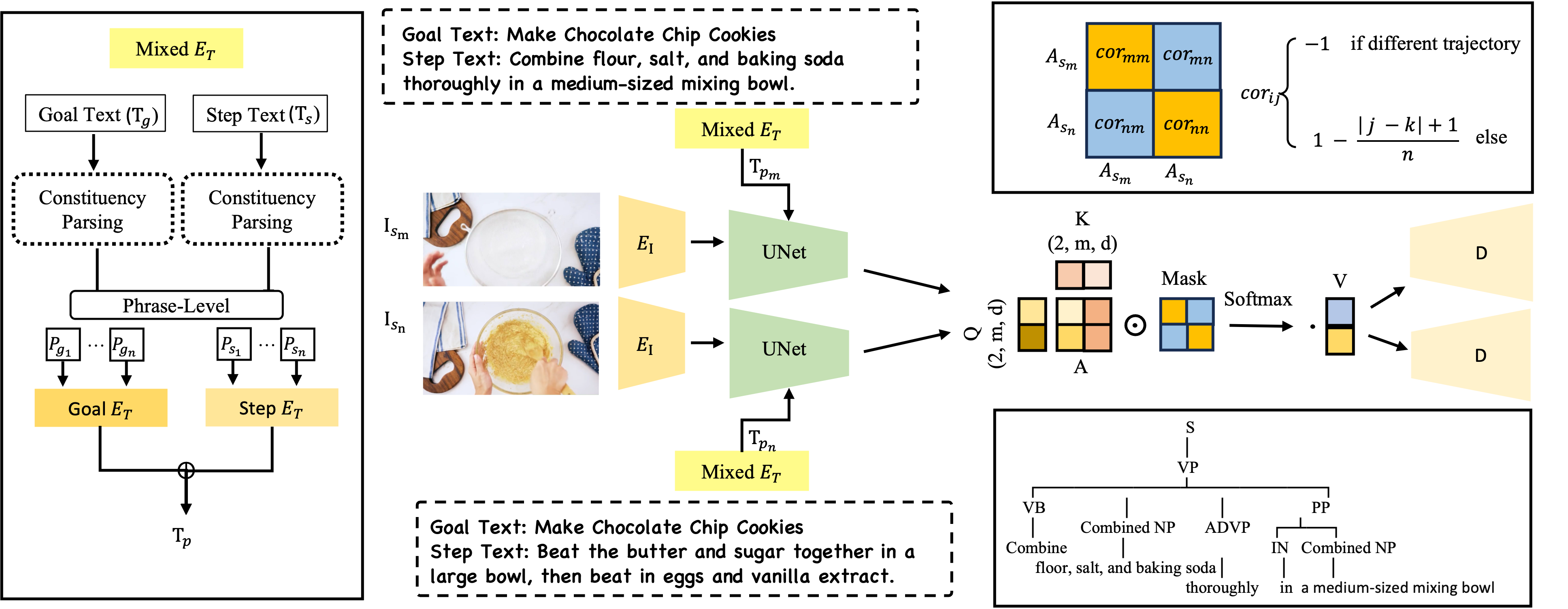}
  \caption{We randomly sample pairs $(v_i,s_i)$ and $(v_j,s_j)$, apply a custom adjacency mask to fuse latent representations, and decode them back into images. The constituency parser (Sec.~\ref{sec:enc}) splits text to handle length constraints.}
  \label{fig:goal-align}
  \vspace{-4mm}
\end{figure*}

\subsection{Effective Text Encoding}
\label{sec:enc}
Our instructional dataset often contains long, detailed step descriptions exceeding typical encoder capacity. To retain critical information, we employ a constituency parser to segment each step into coherent clauses (e.g., verb phrases). Each clause is then separately encoded and concatenated. For goal text (usually short and high-level), we adopt CLIP-ViT/L~\cite{radford2021learning}, while steps are encoded by OpenCLIP-ViT/G~\cite{cherti2023reproducible}, which better handles extended text. This division preserves essential details—particularly when multiple objects or sub-steps appear in a single instruction—minimizing semantic loss.

\subsection{Preference Optimization}
\label{sec:align}
Our model conditions on $(g,s)$ sampled from a pretrained LLM and aims to match ground-truth images more closely. Let $c=(g,s)$ and $x_T\!\sim\!\mathcal{N}(0,I)$:
\[
J(\theta)\;=\;\mathbb{E}_{c,x_T}\Bigl[r\bigl(\text{sample}(\theta,g,s,x_T),\,g,s\bigr)\Bigr],
\]
where $r(\cdot)$ quantifies the alignment of the image-text. We freeze the LLM and update only diffusion parameters $\phi$ via gradient-based feedback through the sampling steps, improving fidelity without retraining the language backbone.

\subsection{Training and Optimization}
We iteratively ascend the gradient of $J(\theta)$, passing reward signals backward through each diffusion step. Over multiple epochs, the model aligns generated visuals more tightly with textual goals/steps. By decoupling text generation from our image diffuser, we can leverage large language models directly, focusing on refining visual accuracy and coherence for complex, multi-step tasks.

\section{Experiment}

\subsection{Datasets}

\noindent \textbf{CaptainCook4D:}
The CaptainCook4D dataset comprises 384 cooking videos, covering 24 cooking tasks. According to the creator of the dataset, the primary objective of creating the dataset was to understand the errors in the procedural activities, so that some of the candidates' operations in the videos did not align with the steps in the instructions. To address this issue, we eliminated the error steps based on the annotation labels.

\noindent \textbf{HT-Step:}
HT-Step is a substantial dataset featuring temporal annotations of instructional steps in cooking videos, derived from the HowTo100M dataset. It encompasses 116k segment-level annotations across 20k narrated videos, totaling 401 unique tasks. HT-Step offers an unprecedented scale and diversity in terms of tasks and the richness of natural language descriptions of steps, providing a robust foundation for aligning instructional articles with how-to videos.

\noindent \textbf{WikiAll:}
Another significant wellspring of instructional visual datasets stems from how-to articles. We meticulously scoured prior research that harnesses these datasets, methodically merging and filtering three distinct sources: VGSI~\cite{yang2021visual}, as well as two other datasets derived from WikiHow, as detailed in papers~\cite{wu2024understanding, 9879568}, to culminate in a synthetic dataset termed WikiAll. This consolidated dataset comprises 87,651 tasks.

\begin{figure*}[h]
  \centering
  \includegraphics[width=1.0\linewidth]{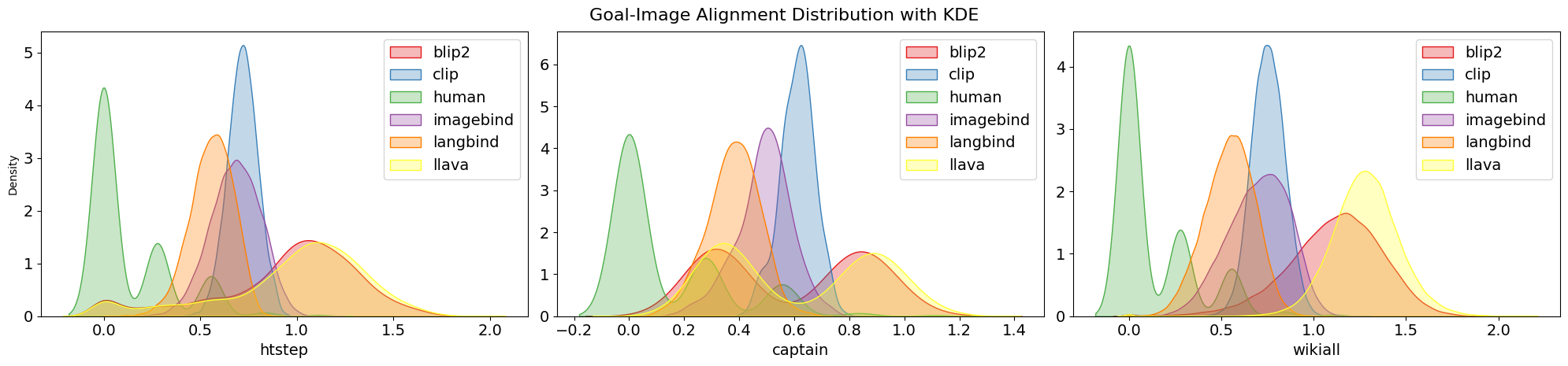}
  \caption{We demonstrate that the goal text often serves as contextual information with limited relation to the visual content, and CLIP frequently produces the highest scores across various datasets. Additionally, the MLLM often fails to align with human judgment, providing high scores that do not correspond with the intended goals.}
  \label{fig:goal-align}
\end{figure*}

\subsection{Text-Image Alignment}
Traditional diffusion model evaluations prioritize image quality, assessing whether generated images resemble real photographs or exhibit aesthetic appeal. However, in instructional visual modeling, the alignment of text with images is crucial. This alignment is typically assessed using tools like CLIPScore~\cite{hessel2021clipscore} due to their simplicity and cost-effectiveness. CLIPScore quantifies semantic accuracy by computing the cosine similarity between the text and the corresponding image and times $w$ to enlarge the differentiation between distribution, where we take $w=2.5$ be consistent. Despite its utility, CLIPScore often assigns low scores in instructional contexts when either goal or step text alignment is inadequate, as illustrated in Fig.~\ref{fig:goal-align}. This recurring issue is likely due to instructional texts being composed of complex natural language, which does not align well with CLIP's capabilities, challenging its effectiveness in such scenarios.

To address the limitations of standard methods like CLIP, we first developed a human evaluation procedure where evaluators assess the alignment between images and goal/step texts. We maintained the same score scale as CLIPScore, where 2.5 indicates high alignment and 0 denotes no relevance.

Given that the dataset is intended to illustrate task execution, we anticipated a strong correlation between the visual information and the associated texts on goals and steps. However, our results reveal that evaluators often struggled to identify clear connections between the goal texts and the visual content. This issue may arise because the goal texts tend to provide contextual rather than explicit visual references. Moreover, although evaluators were instructed to assign higher scores for strong perceived alignment, the scores did not reach the upper end of the scale. This suggests that while there is generally better alignment between step texts and images than between goal texts and images, there is still a lack of consensus among evaluators about what constitutes maximal alignment. Consequently, relying solely on goal-text-image alignment for evaluation may be inadequate, as this metric often demands additional reasoning and imagination that current models do not capture.

Building on these insights, we further evaluated two popular methods: Imagebind and Langbind. Although these occasionally outperformed CLIP in certain datasets (e.g., Captain dataset), they also showed weaknesses, such as lower image-step text alignment scores in the same dataset, as shown in Fig.~\ref{fig:goal-align}.

In search of a more effective approach, we leveraged a caption model to generate image captions, subsequently comparing these captions with the original texts. We employed two widely-used MLLMs, BLIP2 and LLAVA, for captioning, detailed in the appendix. Furthermore, we introduced a text encoder, referenced as ~\cite{behnamghader2024llm2vec}, using a decoder-only model as an encoder to assess the similarity of long texts.

Our findings reveal that this approach aligns well with human judgment across three datasets, particularly in terms of step-text alignment. Notably, in the Htstep dataset, the score histograms varied, but the fitted Kernel Density Estimates (KDE) were nearly identical, indicating strong agreement both with human evaluators and between models. By integrating MLLMs and the text encoder, we established a robust evaluation framework that serves not only as an evaluator but also as a reward model to assist in fine-tuning, discussed further in Sec \ref{sec:align}.

\subsection{Evaluation}

Both "Goal Faithfulness" and "Step Faithfulness" use CLIP to measure how well an image aligns with its intended text. They compare the CLIP similarity of the image with the correct text (either the goal or the step text) against texts from other goals or steps.
Both metrics rely on text-image alignment scores compared with other random steps, using relative CLIP scores as the standard. However, previous sections show that CLIP scores poorly correlate with human preferences.
 
Initial experiments also revealed that the "Cross-Image Consistency" metric in previous work struggles with diffusion model text prompts that fail to differentiate between steps, often generating very similar images for different steps as shown in Figure\ref{fig:instruct}
Essentially, our goal is to model the distribution discrepancy. In our work, we use the Kullback-Leibler (KL) divergence and the Chi-square test as metrics to quantify this discrepancy.
This approach addresses shortcomings observed in previously proposed metrics, enhancing the accuracy of text-image alignment evaluations to better align with human preferences. We choose the below baseline model:

\noindent \textbf{Stable Diffusion 1.5 (SD15) ~\cite{rombach2022high}} SD15 is a text-to-image diffusion model that produces photorealistic images from textual descriptions. The 1.5 version has been fine-tuned through 595k steps at a 512x512 resolution using the "laion-aesthetics v2 5+" dataset, with a 10\% reduction in text-conditioning to enhance classifier-free guidance sampling.

\noindent \textbf{Stable Diffusion 2.1 Turbo (SD21) ~\cite{rombach2022high}} SD21 Turbo is a distillation of SD2.1, itself an iteration of SD2.0, trained from scratch over 550k steps at 256x256 resolution on a subset of LAION-5B, yielding improved results compared to SD1.5.

\begin{figure*}[h]
  \centering
  \includegraphics[width=1.0\linewidth]{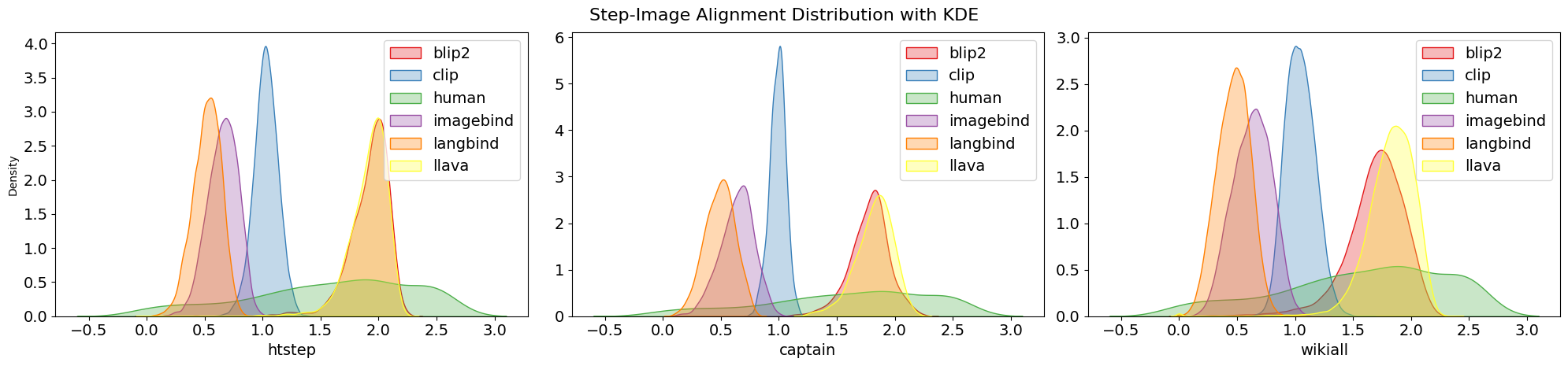}
  \caption{Compared to goal image alignment, MLLMs perform better in aligning step text with images, showing high agreement with human judgment. Although human evaluators tend to assign a range of scores, they generally award high scores.}
  \label{fig:step-align}
\end{figure*}

\begin{table*}[t]
\centering
\caption{Comparison of text-image alignment distribution across procedural language datasets}
\label{tab:qa_results}
\scalebox{0.65}{
    \begin{tabular}{l|l|l|c|c|c|c|c|c|c|c|c}
    \toprule
    \textbf{Method} & \textbf{GT} & \textbf{Captioner} & \multicolumn{3}{|c|}{\textbf{HTStep}} & \multicolumn{3}{|c|}{\textbf{Captain4D}} & \multicolumn{3}{c}{\textbf{Wikiall}} \\
    \midrule
    & & & KL & Chi2 & FID& KL  & Chi2 & FID& KL & Chi2& FID \\
    \midrule
    SD1.5 & Human & llava & 13.01 & 6.67 &53.50 &13.46 & 7.09 &57.09 &14.55 & 6.21 &51.99\\ 
    SD1.5 & Human & blip2 & 12.81 & 6.23 & &12.55 & 7.00 & &14.35 & 6.18 \\ 
    SD2.1 & Human & llava &  14.06 & 6.67 & 47.31&13.77 & 6.91 & 49.24&14.23 & 6.23  & 45.78\\ 
    SD2.1 & Human & blip2 &  13.81 & 6.68 & &13.45 & 6.93 & &14.64 & 6.15 \\ 
    StackedDiffusion & Human & llava & 13.01 & 6.67 &39.65 &12.74 & 7.05 &40.21 &14.55 & 6.22 & 37.72\\ 
    StackedDiffusion & Human & blip2 & 12.91 & 6.56 & &11.72 & 7.87 & &14.12 & 6.19 \\ 
    SDXL & Human & llava & 13.90 & 6.20 &33.23  &12.92 & 6.41 & 35.54&13.28 & 5.20& 36.67\\
    SDXL & Human & blip2 & 13.02 & 6.43 & &14.08& 6.20& &13.95& 5.18\\
    Ours & Human & llava & \textbf{12.71} & \textbf{5.92} &31.87 &\textbf{11.92} & \textbf{4.71} & 34.04& \textbf{12.18} & \textbf{4.70}&32.49\\
    Ours & Human & blip2 & \textbf{12.48} & \textbf{5.22} & &\textbf{10.62} & \textbf{4.91} & &\textbf{11.79} & \textbf{4.43}\\
    \bottomrule
    \end{tabular}
}
\caption*{Note: Lower values indicate better alignment between procedural language and generated visual representation ($\times 100$ for clarity). The FID column provides an image quality baseline across datasets, independent of the language-vision alignment measures.}
\end{table*}

\noindent \textbf{Stable Diffusion xl Turbo(SDXl) ~\cite{podell2023sdxl}} SDXL is the latest and most powerful open source diffusion model in the Stable Diffusion series, featuring a UNet that is three times larger and incorporating a second text encoder (OpenCLIP ViT-bigG/14) alongside the original, significantly increasing its parameter count.

\noindent \textbf{StackedDiffusion~\cite{menon2023generating}} This work is closely related to our approach of tiling along a single spatial dimension and using diffusion latent representations to generate instruction sequences.

From the table, it is clear that our method outperformed the baseline model in both the Chi-squared (CHI2) and KL divergence tests. However, we also observed that although the images generated by the baseline model are not as convincing as illustrations, they received high scores when judged by the MLLM. Upon reviewing the captions generated, we found that the MLLM tends to assign high scores when the object names in the text match, even if the actions described differ. This indicates potential areas for further improvement.

\section*{Conclusion and Limitations}

In this work, we introduced a text-conditioned diffusion framework that decomposes instructional content into goals and step-by-step instructions. Our method leverages a large language model for textual representation, combined with a feedback mechanism for image-text alignment, to generate coherent illustrations of procedural activities. The experimental results demonstrate the effectiveness of our model over baselines in multiple datasets, supported by a comprehensive evaluation protocol.

However, several challenges remain. First, our approach struggles with abstract or implicit actions that do not manifest themselves as clear visual features. Second, while we ensure per-step alignment, uniformity across consecutive images can still be improved, especially for extended multi-step processes. Lastly, integrating multi-turn LLM reasoning into the generative process is a promising avenue for capturing more nuanced instructions and facilitating broader applicability. We will explore these directions in future work.



\bibliography{reference}

\appendix

\section{Appendix}
\label{sec:appendix}

\subsection{Prompt for Generating Image}

\textbf{Goal: Make Classic Chocolate Chip Cookies}

\textbf{Step1:} Sift together flour, baking soda, and salt.

\textbf{Step2:} Beat in eggs and vanilla extract into the butter mixture.

\textbf{Step3:} Gradually mix the dry ingredients into the wet mixture.

\textbf{Step4:} Stir in chocolate chips.

\textbf{Step5:} Chill the dough for 30 minutes in the refrigerator.

\textbf{Step6:} Drop dough by rounded spoonfuls onto baking sheets.

\textbf{Step7:} Bake and let cool on a wire rack.
\\\\
\textbf{Goal: Make Pupusas (Salvadoran Stuffed Corn Cakes)}

\textbf{Step1:} Combine 2 cups of masa harina (corn flour) with about 1.5 cups of warm water and a pinch of salt. Mix until the dough is smooth and pliable. If it's too dry, add more water; if too sticky, add more masa harina.

\textbf{Step2:} Common fillings include shredded cheese, refried beans, and cooked and minced pork (chicharrón). You can mix these fillings together or use them separately.

\textbf{Step3:} Divide the dough into 8-10 equal portions and roll them into balls.

\textbf{Step4:} Press each ball into a flat round disc, making a well in the center for the filling. Add your chosen filling, then fold the dough over the filling and seal it by pressing the edges together.

\textbf{Step5:} Gently flatten the filled balls into thick discs, being careful not to let the filling escape.

\textbf{Step6:} Heat a skillet or griddle over medium-high heat. Cook each pupusa for about 4-5 minutes on each side until they are golden brown and the surface has some charred spots.

\textbf{Step7:} Serve hot with curtido (a spicy cabbage slaw) and tomato salsa.
\\\\
\textbf{Goal: Make Kolaches (Czech Pastries)}

\textbf{Step1:} Dissolve 1 packet of active dry yeast in 1/4 cup of warm water with a teaspoon of sugar. Let it sit until it becomes frothy, about 5-10 minutes.

\textbf{Step2:} In a large bowl, mix 4 cups of flour, 1/4 cup of sugar, and a pinch of salt. In a separate bowl, beat 2 eggs with 1 cup of warm milk and 1/4 cup of melted butter. Combine the yeast mixture with the egg mixture, then gradually add to the dry ingredients to form a dough.

\textbf{Step3:} Turn the dough onto a floured surface and knead until smooth and elastic, about 8-10 minutes. Place in a greased bowl, cover, and let it rise in a warm place until doubled in size, about 1 hour.

\textbf{Step4:} Fillings can vary from sweet (like fruit preserves or sweetened cream cheese) to savory (like sausage or cheese). Prepare your chosen filling.

\textbf{Step5:} Punch down the dough and divide it into about 24 small balls. Flatten each ball slightly, and then press a deep indent in the center. Fill the indent with your filling.

\textbf{Step6:} Arrange the filled dough balls on a baking sheet, cover, and let rise for another 30 minutes.

\textbf{Step7:} Preheat your oven to 375°F (190°C). Bake the kolaches for 15-20 minutes or until golden brown.
\\\\
\textbf{Goal: Make Grilled Steak}

\textbf{Step1:} Rub the raw steak with salt, pepper, and garlic for seasoning.

\textbf{Step2:} Heat the grill to a high temperature of 450°F (232°C).

\textbf{Step3:} Sear the seasoned steak on the grill, flipping once, until it develops a rich, golden crust, about 4-5 minutes per side.

\textbf{Step4:} During the final minutes, baste the steak with a mixture of melted butter and aromatic herbs such as thyme and rosemary.

\textbf{Step5:} Transfer the steak from the grill to a plate and allow it to rest, enhancing its juiciness.

\textbf{Step6:} Cut the steak against the grain into thin slices, revealing the tender, cooked interior.

\textbf{Step7:} Plate the sliced steak with your chosen sides, serving it hot and ready to enjoy.

\subsection{Prompt for Generating Caption}

\textbf{Choice1:} "Describe this image <image> in a detailed manner"

\textbf{Choice2:} "What happened in the picture <image>? Answer in short sentences."

\textbf{Choice3:} "Briefly say the content of this scene <image>"

\textbf{Choice4:} "Show the content in the photo <image> in short text."

\textbf{Choice5:} "Please describe the content of the image <image> in a few words."

\textbf{Choice6:} "What is the content of the image <image>? Please answer in short sentences."

\textbf{Choice7:} "Can you give me a brief description of this image <image>?"

\textbf{Choice8:} "What do you see in this picture <image>?"

\textbf{Choice9:} "In a few words, describe the content of the image <image>."

\textbf{Choice10:} "Provide a concise explanation of this photograph <image>."

\textbf{Choice11:} "What is happening in this scene <image>?"

\textbf{Choice12:} "Summarize the content of the photo <image>."

\textbf{Choice13:} "What are the main elements present in the image <image>?"

\textbf{Choice14:} "Quickly explain the content of this visual <image>."

\textbf{Choice15:} "In a nutshell, what can you say about this picture <image>?"

\textbf{Choice16:} "What's the main subject in the image <image>?"

\textbf{Choice17:} "Describe the main features of the image <image>."

\textbf{Choice18:} "What is depicted in this photograph <image>?"

\textbf{Choice19:} "Give me a short description of the picture <image>."

\textbf{Choice20:} "Briefly describe the objects and actions in the image <image>."

\textbf{Choice21:} "What is the context of this image <image>?"

(Randomly choose one from the above when generating captions.)

\subsection{Successful and Failed Cases}

\begin{figure}[h]
  \centering
  \includegraphics[width=1.05\textwidth]{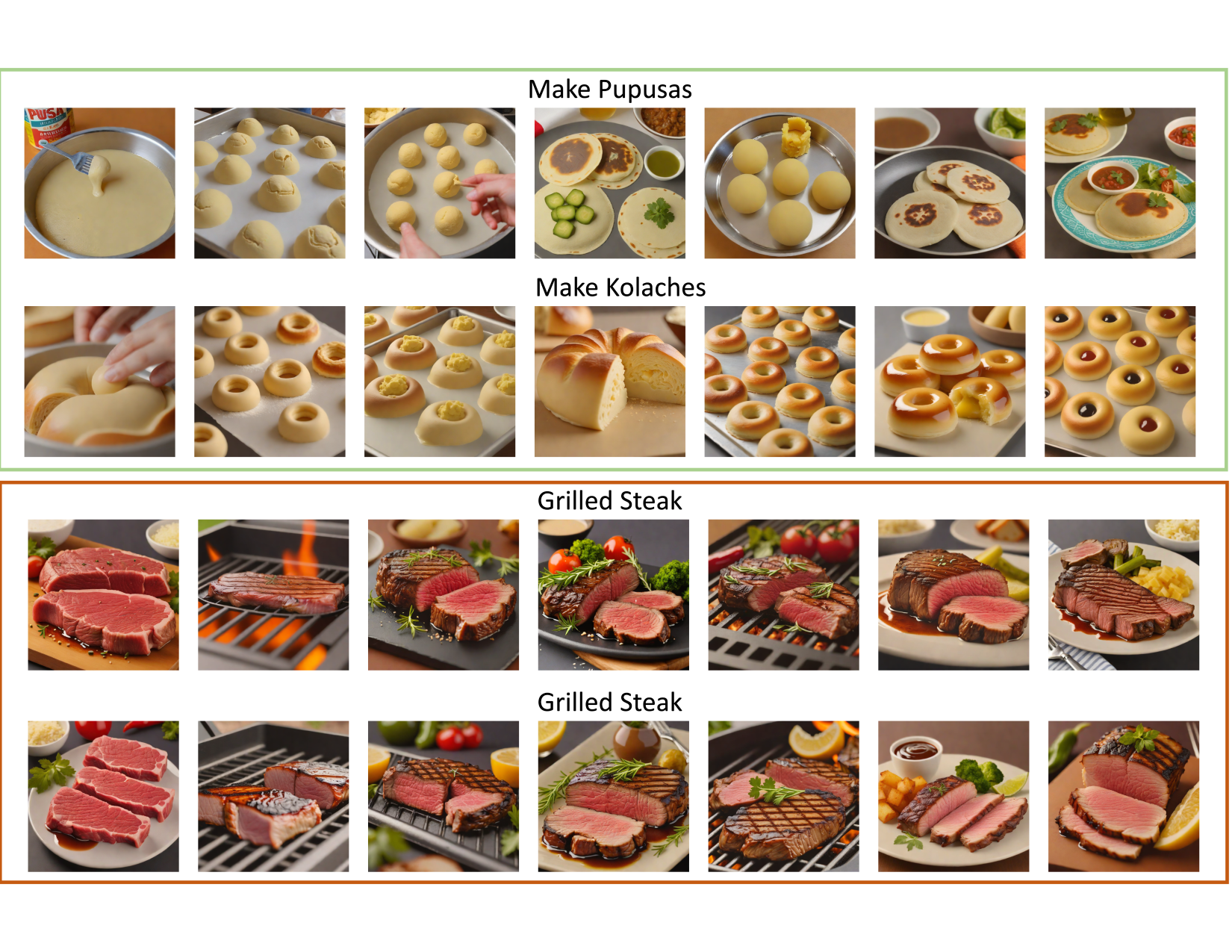}
  \caption{Images of successful and failed generation}
  \label{fig:goal-align}
\end{figure}

We also observed a biased situation where the presence of certain words consistently triggers similar figures, regardless of how much additional instructional text we include in the goal or step descriptions. For example, as shown in Figure \ref{fig:goal-align}, when the word 'steak' appears, the figure invariably displays an angel, failing to represent the progression. This occurs despite explicitly mentioning a stove in the prompt, yet the stove does not appear in the image.


\end{document}

%% file: packages.tex
\usepackage{graphicx}

\usepackage{float} 

%% file: acl_latex.bbl
\begin{thebibliography}{59}
\providecommand{\natexlab}[1]{#1}

\bibitem[{Afouras et~al.(2023)Afouras, Mavroudi, Nagarajan, Wang, and Torresani}]{afouras2023htstep}
Triantafyllos Afouras, Effrosyni Mavroudi, Tushar Nagarajan, Huiyu Wang, and Lorenzo Torresani. 2023.
\newblock \href {https://openreview.net/forum?id=vv3cocNsEK} {{HT}-step: Aligning instructional articles with how-to videos}.
\newblock In \emph{Thirty-seventh Conference on Neural Information Processing Systems Datasets and Benchmarks Track}.

\bibitem[{Bain et~al.(2022)Bain, Nagrani, Varol, and Zisserman}]{bain2022frozen}
Max Bain, Arsha Nagrani, Gül Varol, and Andrew Zisserman. 2022.
\newblock \href {https://arxiv.org/abs/2104.00650} {Frozen in time: A joint video and image encoder for end-to-end retrieval}.
\newblock \emph{Preprint}, arXiv:2104.00650.

\bibitem[{BehnamGhader et~al.(2024)BehnamGhader, Adlakha, Mosbach, Bahdanau, Chapados, and Reddy}]{behnamghader2024llm2vec}
Parishad BehnamGhader, Vaibhav Adlakha, Marius Mosbach, Dzmitry Bahdanau, Nicolas Chapados, and Siva Reddy. 2024.
\newblock \href {https://arxiv.org/abs/2404.05961} {Llm2vec: Large language models are secretly powerful text encoders}.
\newblock \emph{Preprint}, arXiv:2404.05961.

\bibitem[{Betker et~al.()Betker, Goh, Jing, TimBrooks, Wang, Li, LongOuyang, JuntangZhuang, JoyceLee, YufeiGuo, WesamManassra, PrafullaDhariwal, CaseyChu, YunxinJiao, and Ramesh}]{BetkerImprovingIG}
James Betker, Gabriel Goh, Li~Jing, † TimBrooks, Jianfeng Wang, Linjie Li, † LongOuyang, † JuntangZhuang, † JoyceLee, † YufeiGuo, † WesamManassra, † PrafullaDhariwal, † CaseyChu, † YunxinJiao, and Aditya Ramesh.
\newblock \href {https://api.semanticscholar.org/CorpusID:264403242} {Improving image generation with better captions}.

\bibitem[{Black et~al.(2024)Black, Janner, Du, Kostrikov, and Levine}]{black2024training}
Kevin Black, Michael Janner, Yilun Du, Ilya Kostrikov, and Sergey Levine. 2024.
\newblock \href {https://arxiv.org/abs/2305.13301} {Training diffusion models with reinforcement learning}.
\newblock \emph{Preprint}, arXiv:2305.13301.

\bibitem[{Chang et~al.(2023)Chang, Koulieris, and Shum}]{chang2023design}
Ziyi Chang, George~Alex Koulieris, and Hubert P.~H. Shum. 2023.
\newblock \href {https://arxiv.org/abs/2306.04542} {On the design fundamentals of diffusion models: A survey}.
\newblock \emph{Preprint}, arXiv:2306.04542.

\bibitem[{Chen et~al.(2024{\natexlab{a}})Chen, Chen, Zhang, Liu, Wang, Zhou, Zhang, Zhou, Wan, and Sun}]{chen2024mllmasajudge}
Dongping Chen, Ruoxi Chen, Shilin Zhang, Yinuo Liu, Yaochen Wang, Huichi Zhou, Qihui Zhang, Pan Zhou, Yao Wan, and Lichao Sun. 2024{\natexlab{a}}.
\newblock \href {https://arxiv.org/abs/2402.04788} {Mllm-as-a-judge: Assessing multimodal llm-as-a-judge with vision-language benchmark}.
\newblock \emph{Preprint}, arXiv:2402.04788.

\bibitem[{Chen et~al.(2024{\natexlab{b}})Chen, Liu, Yi, Xu, Lai, Wang, Ho, and Xu}]{chen2024evaluating}
Muxi Chen, Yi~Liu, Jian Yi, Changran Xu, Qiuxia Lai, Hongliang Wang, Tsung-Yi Ho, and Qiang Xu. 2024{\natexlab{b}}.
\newblock Evaluating text-to-image generative models: An empirical study on human image synthesis.
\newblock \emph{arXiv preprint arXiv:2403.05125}.

\bibitem[{Chen et~al.(2023)Chen, Liu, and Ding}]{chen2023xiqe}
Yixiong Chen, Li~Liu, and Chris Ding. 2023.
\newblock \href {https://arxiv.org/abs/2305.10843} {X-iqe: explainable image quality evaluation for text-to-image generation with visual large language models}.
\newblock \emph{Preprint}, arXiv:2305.10843.

\bibitem[{Cherti et~al.(2023)Cherti, Beaumont, Wightman, Wortsman, Ilharco, Gordon, Schuhmann, Schmidt, and Jitsev}]{cherti2023reproducible}
Mehdi Cherti, Romain Beaumont, Ross Wightman, Mitchell Wortsman, Gabriel Ilharco, Cade Gordon, Christoph Schuhmann, Ludwig Schmidt, and Jenia Jitsev. 2023.
\newblock Reproducible scaling laws for contrastive language-image learning.
\newblock In \emph{Proceedings of the IEEE/CVF Conference on Computer Vision and Pattern Recognition}, pages 2818--2829.

\bibitem[{Clark et~al.(2023)Clark, Vicol, Swersky, and Fleet}]{clark2023directly}
Kevin Clark, Paul Vicol, Kevin Swersky, and David~J Fleet. 2023.
\newblock \href {https://arxiv.org/abs/2309.17400} {Directly fine-tuning diffusion models on differentiable rewards}.
\newblock \emph{Preprint}, arXiv:2309.17400.

\bibitem[{Dai et~al.(2023)Dai, Hou, Ma, Tsai, Wang, Wang, Zhang, Vandenhende, Wang, Dubey, Yu, Kadian, Radenovic, Mahajan, Li, Zhao, Petrovic, Singh, Motwani, Wen, Song, Sumbaly, Ramanathan, He, Vajda, and Parikh}]{dai2023emu}
Xiaoliang Dai, Ji~Hou, Chih-Yao Ma, Sam Tsai, Jialiang Wang, Rui Wang, Peizhao Zhang, Simon Vandenhende, Xiaofang Wang, Abhimanyu Dubey, Matthew Yu, Abhishek Kadian, Filip Radenovic, Dhruv Mahajan, Kunpeng Li, Yue Zhao, Vladan Petrovic, Mitesh~Kumar Singh, Simran Motwani, and 7 others. 2023.
\newblock \href {https://arxiv.org/abs/2309.15807} {Emu: Enhancing image generation models using photogenic needles in a haystack}.
\newblock \emph{Preprint}, arXiv:2309.15807.

\bibitem[{Dong et~al.(2023)Dong, Xiong, Goyal, Zhang, Chow, Pan, Diao, Zhang, Shum, and Zhang}]{dong2023raft}
Hanze Dong, Wei Xiong, Deepanshu Goyal, Yihan Zhang, Winnie Chow, Rui Pan, Shizhe Diao, Jipeng Zhang, Kashun Shum, and Tong Zhang. 2023.
\newblock \href {https://arxiv.org/abs/2304.06767} {Raft: Reward ranked finetuning for generative foundation model alignment}.
\newblock \emph{Preprint}, arXiv:2304.06767.

\bibitem[{Fan et~al.(2023)Fan, Watkins, Du, Liu, Ryu, Boutilier, Abbeel, Ghavamzadeh, Lee, and Lee}]{fan2023dpok}
Ying Fan, Olivia Watkins, Yuqing Du, Hao Liu, Moonkyung Ryu, Craig Boutilier, Pieter Abbeel, Mohammad Ghavamzadeh, Kangwook Lee, and Kimin Lee. 2023.
\newblock \href {https://arxiv.org/abs/2305.16381} {Dpok: Reinforcement learning for fine-tuning text-to-image diffusion models}.
\newblock \emph{Preprint}, arXiv:2305.16381.

\bibitem[{Fried et~al.(2020)Fried, Alayrac, Blunsom, Dyer, Clark, and Nematzadeh}]{actionseg}
Daniel Fried, Jean-Baptiste Alayrac, Phil Blunsom, Chris Dyer, Stephen Clark, and Aida Nematzadeh. 2020.
\newblock \href {https://arxiv.org/abs/2005.03684} {Learning to segment actions from observation and narration}.
\newblock \emph{Preprint}, arXiv:2005.03684.

\bibitem[{Fu et~al.(2023)Fu, Ng, Jiang, and Liu}]{fu2023gptscore}
Jinlan Fu, See-Kiong Ng, Zhengbao Jiang, and Pengfei Liu. 2023.
\newblock Gptscore: Evaluate as you desire.
\newblock \emph{arXiv preprint arXiv:2302.04166}.

\bibitem[{Gal et~al.(2022)Gal, Alaluf, Atzmon, Patashnik, Bermano, Chechik, and Cohen-Or}]{gal2022image}
Rinon Gal, Yuval Alaluf, Yuval Atzmon, Or~Patashnik, Amit~H. Bermano, Gal Chechik, and Daniel Cohen-Or. 2022.
\newblock \href {https://arxiv.org/abs/2208.01618} {An image is worth one word: Personalizing text-to-image generation using textual inversion}.
\newblock \emph{Preprint}, arXiv:2208.01618.

\bibitem[{Ging et~al.(2020)Ging, Zolfaghari, Pirsiavash, and Brox}]{videoretr}
Simon Ging, Mohammadreza Zolfaghari, Hamed Pirsiavash, and Thomas Brox. 2020.
\newblock \href {https://arxiv.org/abs/2011.00597} {Coot: Cooperative hierarchical transformer for video-text representation learning}.
\newblock \emph{Preprint}, arXiv:2011.00597.

\bibitem[{Han et~al.(2022)Han, Xie, and Zisserman}]{han2022temporal}
Tengda Han, Weidi Xie, and Andrew Zisserman. 2022.
\newblock \href {https://arxiv.org/abs/2204.02968} {Temporal alignment networks for long-term video}.
\newblock \emph{Preprint}, arXiv:2204.02968.

\bibitem[{Hao et~al.(2023)Hao, Chi, Dong, and Wei}]{hao2023optimizing}
Yaru Hao, Zewen Chi, Li~Dong, and Furu Wei. 2023.
\newblock \href {https://arxiv.org/abs/2212.09611} {Optimizing prompts for text-to-image generation}.
\newblock \emph{Preprint}, arXiv:2212.09611.

\bibitem[{Hessel et~al.(2021)Hessel, Holtzman, Forbes, Bras, and Choi}]{hessel2021clipscore}
Jack Hessel, Ari Holtzman, Maxwell Forbes, Ronan~Le Bras, and Yejin Choi. 2021.
\newblock Clipscore: A reference-free evaluation metric for image captioning.
\newblock \emph{arXiv preprint arXiv:2104.08718}.

\bibitem[{Ho et~al.(2020)Ho, Jain, and Abbeel}]{ho2020denoising}
Jonathan Ho, Ajay Jain, and Pieter Abbeel. 2020.
\newblock \href {https://arxiv.org/abs/2006.11239} {Denoising diffusion probabilistic models}.
\newblock \emph{Preprint}, arXiv:2006.11239.

\bibitem[{Kirstain et~al.(2023)Kirstain, Polyak, Singer, Matiana, Penna, and Levy}]{kirstain2023pickapic}
Yuval Kirstain, Adam Polyak, Uriel Singer, Shahbuland Matiana, Joe Penna, and Omer Levy. 2023.
\newblock \href {https://arxiv.org/abs/2305.01569} {Pick-a-pic: An open dataset of user preferences for text-to-image generation}.
\newblock \emph{Preprint}, arXiv:2305.01569.

\bibitem[{Kuehne et~al.(2014)Kuehne, Arslan, and Serre}]{Breakfast}
Hilde Kuehne, Ali Arslan, and Thomas Serre. 2014.
\newblock \href {https://doi.org/10.1109/CVPR.2014.105} {The language of actions: Recovering the syntax and semantics of goal-directed human activities}.
\newblock In \emph{2014 IEEE Conference on Computer Vision and Pattern Recognition}, pages 780--787.

\bibitem[{Lee et~al.(2023)Lee, Liu, Ryu, Watkins, Du, Boutilier, Abbeel, Ghavamzadeh, and Gu}]{lee2023aligning}
Kimin Lee, Hao Liu, Moonkyung Ryu, Olivia Watkins, Yuqing Du, Craig Boutilier, Pieter Abbeel, Mohammad Ghavamzadeh, and Shixiang~Shane Gu. 2023.
\newblock \href {https://arxiv.org/abs/2302.12192} {Aligning text-to-image models using human feedback}.
\newblock \emph{Preprint}, arXiv:2302.12192.

\bibitem[{Lee et~al.(2025)Lee, Kang, and Kim}]{lee2025capo}
Kyungmin Lee, Hyeongjin Kang, and Sungwoong Kim. 2025.
\newblock Calibrated multi-preference optimization for aligning diffusion models.
\newblock \emph{arXiv preprint arXiv:2502.02588}.

\bibitem[{Li et~al.(2024)Li, Kallidromitis, Gokul, Kato, and Kozuka}]{Li2024AligningDM}
Shufan Li, Konstantinos Kallidromitis, Akash Gokul, Yusuke Kato, and Kazuki Kozuka. 2024.
\newblock \href {https://api.semanticscholar.org/CorpusID:269004992} {Aligning diffusion models by optimizing human utility}.
\newblock \emph{ArXiv}, abs/2404.04465.

\bibitem[{Lin et~al.(2022)Lin, Petroni, Bertasius, Rohrbach, Chang, and Torresani}]{stepclass}
Xudong Lin, Fabio Petroni, Gedas Bertasius, Marcus Rohrbach, Shih-Fu Chang, and Lorenzo Torresani. 2022.
\newblock \href {https://arxiv.org/abs/2201.10990} {Learning to recognize procedural activities with distant supervision}.
\newblock \emph{Preprint}, arXiv:2201.10990.

\bibitem[{Liu et~al.(2025)Liu, Duan, Yang et~al.}]{liu2025videodpo}
Runtao Liu, Ning Duan, Zhenyang Yang, and 1 others. 2025.
\newblock Videodpo: Omni-preference alignment for video diffusion generation.
\newblock \emph{arXiv preprint arXiv:2412.14167}.
\newblock To appear at CVPR 2025.

\bibitem[{Liu et~al.(2022)Liu, Li, Wu, Chen, Shan, and Qie}]{temporal}
Ye~Liu, Siyuan Li, Yang Wu, Chang~Wen Chen, Ying Shan, and Xiaohu Qie. 2022.
\newblock \href {https://arxiv.org/abs/2203.12745} {Umt: Unified multi-modal transformers for joint video moment retrieval and highlight detection}.
\newblock \emph{Preprint}, arXiv:2203.12745.

\bibitem[{Luo et~al.(2025)Luo, Lambert, Fu et~al.}]{luo2025internetvideo}
Calvin Luo, Jacob Lambert, Eric Fu, and 1 others. 2025.
\newblock Solving new tasks by adapting internet video knowledge.
\newblock \emph{arXiv preprint arXiv:2504.15369}.

\bibitem[{Menon et~al.(2023)Menon, Misra, and Girdhar}]{menon2023generating}
Sachit Menon, Ishan Misra, and Rohit Girdhar. 2023.
\newblock \href {https://arxiv.org/abs/2312.04552} {Generating illustrated instructions}.
\newblock \emph{Preprint}, arXiv:2312.04552.

\bibitem[{Miech et~al.(2020)Miech, Alayrac, Smaira, Laptev, Sivic, and Zisserman}]{miech2020endtoend}
Antoine Miech, Jean-Baptiste Alayrac, Lucas Smaira, Ivan Laptev, Josef Sivic, and Andrew Zisserman. 2020.
\newblock \href {https://arxiv.org/abs/1912.06430} {End-to-end learning of visual representations from uncurated instructional videos}.
\newblock \emph{Preprint}, arXiv:1912.06430.

\bibitem[{Peddi et~al.(2023)Peddi, Arya, Challa, Pallapothula, Vyas, Wang, Zhang, Komaragiri, Ragan, Ruozzi, Xiang, and Gogate}]{peddi2023captaincook4d}
Rohith Peddi, Shivvrat Arya, Bharath Challa, Likhitha Pallapothula, Akshay Vyas, Jikai Wang, Qifan Zhang, Vasundhara Komaragiri, Eric Ragan, Nicholas Ruozzi, Yu~Xiang, and Vibhav Gogate. 2023.
\newblock \href {https://arxiv.org/abs/2312.14556} {Captaincook4d: A dataset for understanding errors in procedural activities}.
\newblock \emph{Preprint}, arXiv:2312.14556.

\bibitem[{Podell et~al.(2023)Podell, English, Lacey, Blattmann, Dockhorn, Müller, Penna, and Rombach}]{podell2023sdxl}
Dustin Podell, Zion English, Kyle Lacey, Andreas Blattmann, Tim Dockhorn, Jonas Müller, Joe Penna, and Robin Rombach. 2023.
\newblock \href {https://arxiv.org/abs/2307.01952} {Sdxl: Improving latent diffusion models for high-resolution image synthesis}.
\newblock \emph{Preprint}, arXiv:2307.01952.

\bibitem[{Prabhudesai et~al.(2023)Prabhudesai, Goyal, Pathak, and Fragkiadaki}]{prabhudesai2023aligning}
Mihir Prabhudesai, Anirudh Goyal, Deepak Pathak, and Katerina Fragkiadaki. 2023.
\newblock \href {https://arxiv.org/abs/2310.03739} {Aligning text-to-image diffusion models with reward backpropagation}.
\newblock \emph{Preprint}, arXiv:2310.03739.

\bibitem[{Radford et~al.(2021)Radford, Kim, Hallacy, Ramesh, Goh, Agarwal, Sastry, Askell, Mishkin, Clark et~al.}]{radford2021learning}
Alec Radford, Jong~Wook Kim, Chris Hallacy, Aditya Ramesh, Gabriel Goh, Sandhini Agarwal, Girish Sastry, Amanda Askell, Pamela Mishkin, Jack Clark, and 1 others. 2021.
\newblock Learning transferable visual models from natural language supervision.
\newblock In \emph{International conference on machine learning}, pages 8748--8763. PMLR.

\bibitem[{Rombach et~al.(2022)Rombach, Blattmann, Lorenz, Esser, and Ommer}]{rombach2022high}
Robin Rombach, Andreas Blattmann, Dominik Lorenz, Patrick Esser, and Bj{\"o}rn Ommer. 2022.
\newblock High-resolution image synthesis with latent diffusion models.
\newblock In \emph{Proceedings of the IEEE/CVF conference on computer vision and pattern recognition}, pages 10684--10695.

\bibitem[{Samel et~al.(2025)Samel, Sontakke, and Essa}]{samel2025pivot}
Karan Samel, Nitish Sontakke, and Irfan Essa. 2025.
\newblock Leveraging procedural knowledge and task hierarchies for efficient instructional video pre-training.
\newblock \emph{arXiv preprint arXiv:2502.17352}.

\bibitem[{Segalis et~al.(2023)Segalis, Valevski, Lumen, Matias, and Leviathan}]{segalis2023picture}
Eyal Segalis, Dani Valevski, Danny Lumen, Yossi Matias, and Yaniv Leviathan. 2023.
\newblock \href {https://arxiv.org/abs/2310.16656} {A picture is worth a thousand words: Principled recaptioning improves image generation}.
\newblock \emph{Preprint}, arXiv:2310.16656.

\bibitem[{Stein and Mckenna(2013)}]{50Salads}
Sebastian Stein and Stephen Mckenna. 2013.
\newblock \href {https://doi.org/10.1145/2506023.2506031} {User-adaptive models for recognizing food preparation activities}.
\newblock pages 39--44.

\bibitem[{Tang et~al.(2019)Tang, Ding, Rao, Zheng, Zhang, Zhao, Lu, and Zhou}]{COIN}
Yansong Tang, Dajun Ding, Yongming Rao, Yu~Zheng, Danyang Zhang, Lili Zhao, Jiwen Lu, and Jie Zhou. 2019.
\newblock \href {https://doi.org/10.1109/CVPR.2019.00130} {Coin: A large-scale dataset for comprehensive instructional video analysis}.
\newblock In \emph{2019 IEEE/CVF Conference on Computer Vision and Pattern Recognition (CVPR)}, pages 1207--1216.

\bibitem[{Wallace et~al.(2023)Wallace, Dang, Rafailov, Zhou, Lou, Purushwalkam, Ermon, Xiong, Joty, and Naik}]{wallace2023diffusion}
Bram Wallace, Meihua Dang, Rafael Rafailov, Linqi Zhou, Aaron Lou, Senthil Purushwalkam, Stefano Ermon, Caiming Xiong, Shafiq Joty, and Nikhil Naik. 2023.
\newblock \href {https://arxiv.org/abs/2311.12908} {Diffusion model alignment using direct preference optimization}.
\newblock \emph{Preprint}, arXiv:2311.12908.

\bibitem[{Wang et~al.(2023{\natexlab{a}})Wang, Liang, Meng, Shi, Li, Xu, Qu, and Zhou}]{wang2023chatgpt}
Jiaan Wang, Yunlong Liang, Fandong Meng, Haoxiang Shi, Zhixu Li, Jinan Xu, Jianfeng Qu, and Jie Zhou. 2023{\natexlab{a}}.
\newblock Is chatgpt a good nlg evaluator? a preliminary study.
\newblock \emph{arXiv preprint arXiv:2303.04048}.

\bibitem[{Wang et~al.(2025)Wang, Cui, Hou et~al.}]{wang2025cigeval}
Jifang Wang, Rundi Cui, Lei Hou, and 1 others. 2025.
\newblock Cigeval: A unified agentic framework for evaluating conditional image generation.
\newblock \emph{arXiv preprint arXiv:2504.07046}.

\bibitem[{Wang et~al.(2023{\natexlab{b}})Wang, Li, Chen, Zhu, Lin, Cao, Liu, Liu, and Sui}]{wang2023large}
Peiyi Wang, Lei Li, Liang Chen, Dawei Zhu, Binghuai Lin, Yunbo Cao, Qi~Liu, Tianyu Liu, and Zhifang Sui. 2023{\natexlab{b}}.
\newblock Large language models are not fair evaluators.
\newblock \emph{arXiv preprint arXiv:2305.17926}.

\bibitem[{Wei et~al.(2025)Wei, Zhao, Garg, and Shi}]{wei2025instructionbench}
Haiwan Wei, Chen Zhao, Abhishek Garg, and Jianbo Shi. 2025.
\newblock Instructionbench: An instructional video understanding benchmark.
\newblock \emph{arXiv preprint arXiv:2504.05040}.

\bibitem[{Wu et~al.(2025)Wu, Yin, Liu, and He}]{wu2025survey}
Sihao Wu, Kun Yin, Yanhong Liu, and Di~He. 2025.
\newblock Preference alignment on diffusion models: A comprehensive survey for image generation and editing.
\newblock \emph{arXiv preprint arXiv:2502.07829}.

\bibitem[{Wu et~al.(2024)Wu, Spangher, Alipoormolabashi, Freedman, Weischedel, and Peng}]{wu2024understanding}
Te-Lin Wu, Alex Spangher, Pegah Alipoormolabashi, Marjorie Freedman, Ralph Weischedel, and Nanyun Peng. 2024.
\newblock \href {https://arxiv.org/abs/2110.08486} {Understanding multimodal procedural knowledge by sequencing multimodal instructional manuals}.
\newblock \emph{Preprint}, arXiv:2110.08486.

\bibitem[{Wu et~al.(2023)Wu, Hao, Sun, Chen, Zhu, Zhao, and Li}]{wu2023human}
Xiaoshi Wu, Yiming Hao, Keqiang Sun, Yixiong Chen, Feng Zhu, Rui Zhao, and Hongsheng Li. 2023.
\newblock \href {https://arxiv.org/abs/2306.09341} {Human preference score v2: A solid benchmark for evaluating human preferences of text-to-image synthesis}.
\newblock \emph{Preprint}, arXiv:2306.09341.

\bibitem[{Xu et~al.(2023)Xu, Liu, Wu, Tong, Li, Ding, Tang, and Dong}]{xu2023imagereward}
Jiazheng Xu, Xiao Liu, Yuchen Wu, Yuxuan Tong, Qinkai Li, Ming Ding, Jie Tang, and Yuxiao Dong. 2023.
\newblock \href {https://arxiv.org/abs/2304.05977} {Imagereward: Learning and evaluating human preferences for text-to-image generation}.
\newblock \emph{Preprint}, arXiv:2304.05977.

\bibitem[{Yan et~al.(2025)Yan, Liu, Yuan et~al.}]{yan2025gptimgeval}
Zhiyuan Yan, Yutong Liu, Yaqi Yuan, and 1 others. 2025.
\newblock Gpt-imgeval: A benchmark for diagnosing gpt-4v's image generation.
\newblock \emph{arXiv preprint arXiv:2504.02782}.

\bibitem[{Yang et~al.(2025)Yang, Zhao, and Liu}]{yang2025planllm}
Dejie Yang, Zijing Zhao, and Yang Liu. 2025.
\newblock Planllm: Video procedure planning with refinable large language models.
\newblock In \emph{Proceedings of the 39th AAAI Conference on Artificial Intelligence (AAAI)}.

\bibitem[{Yang et~al.(2021{\natexlab{a}})Yang, Bisk, and Gao}]{yang2021taco}
Jianwei Yang, Yonatan Bisk, and Jianfeng Gao. 2021{\natexlab{a}}.
\newblock \href {https://arxiv.org/abs/2108.09980} {Taco: Token-aware cascade contrastive learning for video-text alignment}.
\newblock \emph{Preprint}, arXiv:2108.09980.

\bibitem[{Yang et~al.(2022)Yang, Chen, Jiang, Chen, Wang, and Zhao}]{9879568}
Jinhui Yang, Xianyu Chen, Ming Jiang, Shi Chen, Louis Wang, and Qi~Zhao. 2022.
\newblock \href {https://doi.org/10.1109/CVPR52688.2022.01518} {Visualhow: Multimodal problem solving}.
\newblock In \emph{2022 IEEE/CVF Conference on Computer Vision and Pattern Recognition (CVPR)}, pages 15606--15616.

\bibitem[{Yang et~al.(2023)Yang, Tao, Lyu, Ge, Chen, Li, Shen, Zhu, and Li}]{Yang2023UsingHF}
Kai Yang, Jian Tao, Jiafei Lyu, Chunjiang Ge, Jiaxin Chen, Qimai Li, Weihan Shen, Xiaolong Zhu, and Xiu Li. 2023.
\newblock \href {https://api.semanticscholar.org/CorpusID:265352082} {Using human feedback to fine-tune diffusion models without any reward model}.
\newblock \emph{ArXiv}, abs/2311.13231.

\bibitem[{Yang et~al.(2024)Yang, Zhang, Song, Hong, Xu, Zhao, Zhang, Cui, and Yang}]{yang2024diffusion}
Ling Yang, Zhilong Zhang, Yang Song, Shenda Hong, Runsheng Xu, Yue Zhao, Wentao Zhang, Bin Cui, and Ming-Hsuan Yang. 2024.
\newblock \href {https://arxiv.org/abs/2209.00796} {Diffusion models: A comprehensive survey of methods and applications}.
\newblock \emph{Preprint}, arXiv:2209.00796.

\bibitem[{Yang et~al.(2021{\natexlab{b}})Yang, Panagopoulou, Lyu, Zhang, Yatskar, and Callison-Burch}]{yang2021visual}
Yue Yang, Artemis Panagopoulou, Qing Lyu, Li~Zhang, Mark Yatskar, and Chris Callison-Burch. 2021{\natexlab{b}}.
\newblock Visual goal-step inference using wikihow.
\newblock \emph{arXiv preprint arXiv:2104.05845}.

\bibitem[{Zhukov et~al.(2019)Zhukov, Alayrac, Cinbis, Fouhey, Laptev, and Sivic}]{zhukov2019crosstask}
Dimitri Zhukov, Jean-Baptiste Alayrac, Ramazan~Gokberk Cinbis, David Fouhey, Ivan Laptev, and Josef Sivic. 2019.
\newblock \href {https://arxiv.org/abs/1903.08225} {Cross-task weakly supervised learning from instructional videos}.
\newblock \emph{Preprint}, arXiv:1903.08225.

\end{thebibliography}
